\documentclass[10pt,twocolumn,letterpaper]{article}

\usepackage{cvpr}
\usepackage{times}
\usepackage{graphicx}
\usepackage{amsmath}
\usepackage{amssymb}
\usepackage{booktabs}
\usepackage{multirow}
\usepackage{array}
\usepackage[toc,page,titletoc]{appendix}
\usepackage[pagebackref=true,breaklinks=true,letterpaper=true,colorlinks,bookmarks=false]{hyperref}

\cvprfinalcopy

\newcommand*\samethanks[1][\value{footnote}]{\footnotemark[#1]}

\ifcvprfinal\fi

\begin{document}

\title{Transparency by Design: Closing the Gap Between Performance and Interpretability in Visual
  Reasoning}

\author{David~Mascharka\thanks{Indicates equal contribution.}\hspace{2pt} $^1$ \qquad Philip~Tran$^2$
  \qquad Ryan~Soklaski$^1$ \qquad Arjun~Majumdar\samethanks\hspace{2pt} $^1$\\
  \hfill$^1$MIT~Lincoln~Laboratory\thanks{This material is based upon work supported by the
      Assistant Secretary of Defense for Research and Engineering under Air Force Contract
      No. FA8721-05-C-0002 and/or FA8702-15-D-0001. Any opinions, findings, conclusions or
      recommendations expressed in this material are those of the author(s) and do not necessarily
      reflect the views of the Assistant Secretary of Defense for Research and Engineering.}\hfill
  $^2$Planck Aerosystems\thanks{This work conducted while
    Philip was at MIT Lincoln Laboratory.}\hfill\mbox{ }\\
{\tt\small \{first.last\}@ll.mit.edu, phil@planckaero.com}}

\maketitle

\begin{abstract}
  Visual question answering requires high-order reasoning about an image, which is a fundamental
  capability needed by machine systems to follow complex directives. Recently, modular networks have
  been shown to be an effective framework for performing visual reasoning tasks. While modular
  networks were initially designed with a degree of model transparency, their performance on complex
  visual reasoning benchmarks was lacking. Current state-of-the-art approaches do not provide an
  effective mechanism for understanding the reasoning process. In this paper, we close the
  performance gap between interpretable models and state-of-the-art visual reasoning methods. We
  propose a set of visual-reasoning primitives which, when composed, manifest as a model capable of
  performing complex reasoning tasks in an explicitly-interpretable manner. The fidelity and
  interpretability of the primitives' outputs enable an unparalleled ability to diagnose the
  strengths and weaknesses of the resulting model. Critically, we show that these primitives are
  highly performant, achieving state-of-the-art accuracy of 99.1\% on the CLEVR dataset. We also
  show that our model is able to effectively learn generalized representations when provided a small
  amount of data containing novel object attributes. Using the CoGenT generalization task, we show
  more than a 20 percentage point improvement over the current state of the art.
\end{abstract}

\section{Introduction}\label{sec:introduction}
A visual question answering (VQA) model must be capable of complex spatial reasoning over an image.
For example, in order to answer the question ``What color is the cube to the right of the large
metal sphere?'', a model must identify which sphere is the large metal one, understand what it means
for an object to be to the right of another, and apply this concept spatially to the attended
sphere. Within this new region of interest, the model must find the cube and determine its
color. This behavior should be compositional to allow for arbitrarily long reasoning chains.

\begin{figure}[!t]
  \centering
  \includegraphics[keepaspectratio=true, width=\columnwidth]{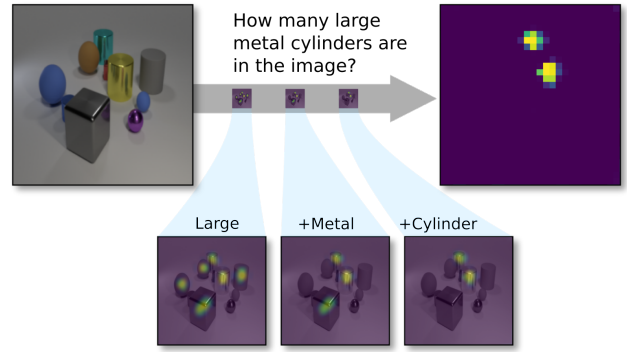}
  \caption{A diagram of a visual question answering task, in which our proposed Transparency by
    Design network (TbD-net) composes a series of attention masks that allow it to correctly count
    two large metal cylinders in the image.}
  \label{fig:tbd-net}
\end{figure}

While a wide variety of models have recently been proposed for the VQA task \cite{mcb, n2nmn,
  no-strong-priors, relational-reasoning, sans, mcb-co}, neural module networks \cite{nmn,
  learning-to-compose, n2nmn, clevr-iep} are among the most intuitive. Introduced by Andreas \etal
\cite{nmn}, neural module networks compose a question-specific neural network, drawing from a set of
modules that each perform an individual operation. This design closely models the compositional
nature of visual reasoning tasks. In the original work, modules were designed with an attention
mechanism, which allowed for insight into the model's operation. However, the approach did not
perform well on complex visual reasoning tasks such as CLEVR \cite{clevr}. Modifications by Johnson
\etal \cite{clevr-iep} address the performance issue at the cost of losing model transparency. This
is problematic, because the ability to inspect each step of the reasoning process is crucial for
real-world applications, in order to ensure proper model behavior, build user trust, and diagnose
errors in reasoning.

Our work closes the gap between performant and interpretable models by designing a module network
explicitly built around a visual attention mechanism. We refer to this approach as Transparency by
Design (TbD), illustrated in Figure~\ref{fig:tbd-net}. As Lipton \cite{mythos-interpretability}
notes, transparency and interpretability are often spoken of but rarely defined. Here, transparency
refers to the ability to examine the intermediate outputs of each module and understand their
behavior at a high level. That is, the module outputs are interpretable if they visually highlight
the correct regions of the input image. This ensures the reasoning process can be interpreted. We
concretely define this notion in Section~\ref{sec:performance}, and provide a quantitative
analysis. In this paper, we:
\begin{enumerate}
  \itemsep0em
  \item Propose a set of composable visual reasoning primitives that incorporate an attention
    mechanism, which allows for model transparency.
  \item Demonstrate state-of-the-art performance on the CLEVR \cite{clevr} dataset.
  \item Show that compositional visual attention provides powerful insight into model behavior.
  \item Propose a method to quantitatively evaluate the interpretability of visual attention
    mechanisms.
  \item Improve upon the current state-of-the-art performance on the CoGenT generalization task
    \cite{clevr} by 20 percentage points.
\end{enumerate}

The structure of this paper is as follows. In Section~\ref{sec:related}, we discuss related work in
visual question answering and visual reasoning, which motivates the incorporation of an explicit
attention mechanism in our model. Section~\ref{sec:model} presents the Transparency by Design
networks. In Section~\ref{sec:experiments}, we present our VQA experiments and results. A discussion
of our contributions is presented in Section~\ref{sec:discussion}. The code for replicating our
experiments is available at \url{https://github.com/davidmascharka/tbd-nets}.

\section{Related Work}\label{sec:related}
Visual question answering (VQA) requires reasoning over both visual and textual information. A
natural-language component must be used to understand the question that is asked, and a visual
component must reason over the provided image in order to answer that question. The two main methods
to address this problem are (1) to parse the question into a series of logical operations, then
perform each operation over the image features or (2) to embed both the image and question into a
feature space, and then reason over the features jointly.

\textbf{Neural Module Networks} (NMNs) follow the first approach. NMNs were introduced by Andreas
\etal \cite{nmn}, and later extended by Andreas \etal \cite{learning-to-compose}, Johnson \etal
\cite{clevr-iep}, and Hu \etal \cite{n2nmn}. A natural-language component parses the given
question and determines the series of logical steps that should be carried out to answer the
question. A \emph{module} is a small neural network used to perform a given logical step. By
composing the appropriate modules, the logical program produced by the natural language component is
carried out and an answer is produced. For example, to answer ``What color is the large metal
cube?'', the output of a module that locates large objects can be composed with a module that finds
things made of metal, then with a module that localizes cubes. A module that determines the color of
objects can then be given the cube module's output to produce an answer.

The original work by Andreas \etal \cite{nmn} provided an attention mechanism, which allowed for a
degree of model transparency. However, their model struggled with long chains of reasoning and
global context. The later work of Andreas \etal \cite{learning-to-compose} focused on improving the
flexibility of the natural-language component and on learning to compose modules rather than dictate
how they should be composed. The modifications by Hu \etal \cite{n2nmn} built off this work,
focusing on incorporating question features into the network modules and improving the
natural-language parser that determines how modules should be composed. While achieving higher
accuracy than its predecessors, this model also struggles with long chains of reasoning and does not
perform as well as other methods on visual reasoning benchmarks such as CLEVR \cite{clevr}.

Johnson \etal \cite{clevr-iep} built on the NMN approach by modifying the natural language component
of their network to allow for more flexibility and developing a set of modules whose generic design
is shared across several operations. These modifications led to an impressive increase in
performance on the CLEVR dataset. However, their modules are not easily interpretable, because they
process high-dimensional features throughout their entire network. The gradient-based mechanism
through which they, along with several others \cite{person-reid, gradcam}, visualize attention can
be limiting.

\begin{figure}[!t]
  \centering
  \includegraphics[keepaspectratio=true, width=0.33\columnwidth]{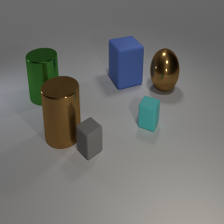}\hfill
  \includegraphics[keepaspectratio=true, width=0.33\columnwidth]{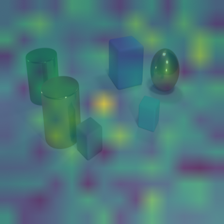}\hfill
  \includegraphics[keepaspectratio=true, width=0.33\columnwidth]{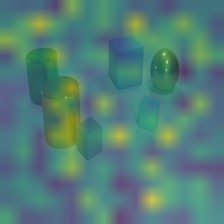}
  \caption{Gradient-based visualizations of an intermediate output (attention on the brown cylinder)
    of a neural module network produce unreliable attention masks. Furthermore, changing a
    downstream module from query color (middle) to query size (right) alters the visualization of
    the attention.}
  \label{fig:different-attentions}
\end{figure}

Gradient-based methods can provide reasonable visualizations at the penultimate layer
\cite{clevr-iep} of a neural module network. However, as depicted in
Figure~\ref{fig:different-attentions}, the regions of attention produced for an intermediate module
are unreliable, and because gradient-based methods flow backward through a network, these
visualizations inappropriately depend on downstream modules in the network.

\textbf{Attention.} Some approaches \cite{bottom-up-top-down, abc-cnn, hierarchical-coattention,
  ask-attend-answer, show-attend-tell, sans} propose an attention mechanism whereby each word
corresponds to some feature of an image. One major difficulty with this type of approach is that
some words have no clear semantic content in image-space. For example, the word `sitting' does not
have a clear region of focus in the question ``What object is the man sitting on?'', and seems to
rely on further analysis of the semantics of the question. The key components of this question are
`man' and `object being [sat] on.' This is a problem for the natural language processing pipeline
rather than the visual component of a system.

Several authors \cite{segmentation-aware, n2nmn, residual-attention} have proposed attention
mechanisms that a network can use, optionally. In the context of providing transparent models, this
can be problematic as a network can learn not to use an attended region at all. By explicitly
forcing the attention mechanism to be used, we ensure our network uses attended regions in an
intuitive way.

Many authors \cite{bottom-up-top-down, abc-cnn, mcb, video-summarization, picanet,
  hierarchical-coattention, ask-attend-answer, show-attend-tell, where-to-look, sans, visual7w} use
a spatial softmax to compute attention weights. This enforces a global normalization across an
image, which results in scene-dependent attention magnitudes. For example, in an image with a single
car, a model asked to attend to the cars would ideally put zero attention on every region that does
not contain a car and full (unity) attention on the region containing the car. In an image with two
cars, a spatial softmax will force each car region to have an attention magnitude of one-half. This
issue is noted by Zhang \etal \cite{zhang2018learning} in the context of counting, but we note a
more general problem. To addres this, we utilize an elementwise sigmoid to ensure that the
activation at each pixel lies between zero and one, and do not introduce any form of global
normalization. Further details on our network architecture and motivation are supplied in the
following section.

\section{Transparency by Design}\label{sec:model}
Breaking a complex chain of reasoning into a series of smaller subproblems, each of which can be
solved independently and composed, is a powerful and intuitive means for reasoning. This type of
modular structure also permits inspection of the the network output at each step in the reasoning
process, contingent on module designs that produce interpretable outputs. Motivated by this, we
introduce a neural module network that explicitly models an attention mechanism in image space,
which we call a Transparency by Design network (TbD-net), following the fact that transparency is a
motivating factor in our design decisions. Meant to achieve performance at the level of the model
from Johnson \etal \cite{clevr-iep} while providing transparency similar to Andreas \etal \cite{nmn}
and Hu \etal \cite{n2nmn}, our model incorporates design decisions from all three architectures. The
program generator from Johnson \etal \cite{clevr-iep} allows for impressive flexibility and performs
exceptionally well, so we reuse this component in our network. We thus use their set of primitive
operations, listed in Table~\ref{tab:modules}, but redesign each module according to its intended
function. The resulting modules are similar in spirit to the approaches taken by Andreas \etal
\cite{nmn} and Hu \etal \cite{n2nmn}.

To motivate this design decision, consider that some modules need only focus on local features in an
image, as in the case of an \texttt{Attention} module which focuses on distinct objects or
properties. Other modules need global context in order to carry out their operation, as in the case
of \texttt{Relate} modules, which must be capable of shifting attention across an entire image. We
combine our prior knowledge about each module's task with empirical experimentation, resulting in a
set of novel module architectures optimized for each operation.

\begin{figure}[!ht]
  \centering
  \includegraphics[keepaspectratio=true, height=0.75\textheight]{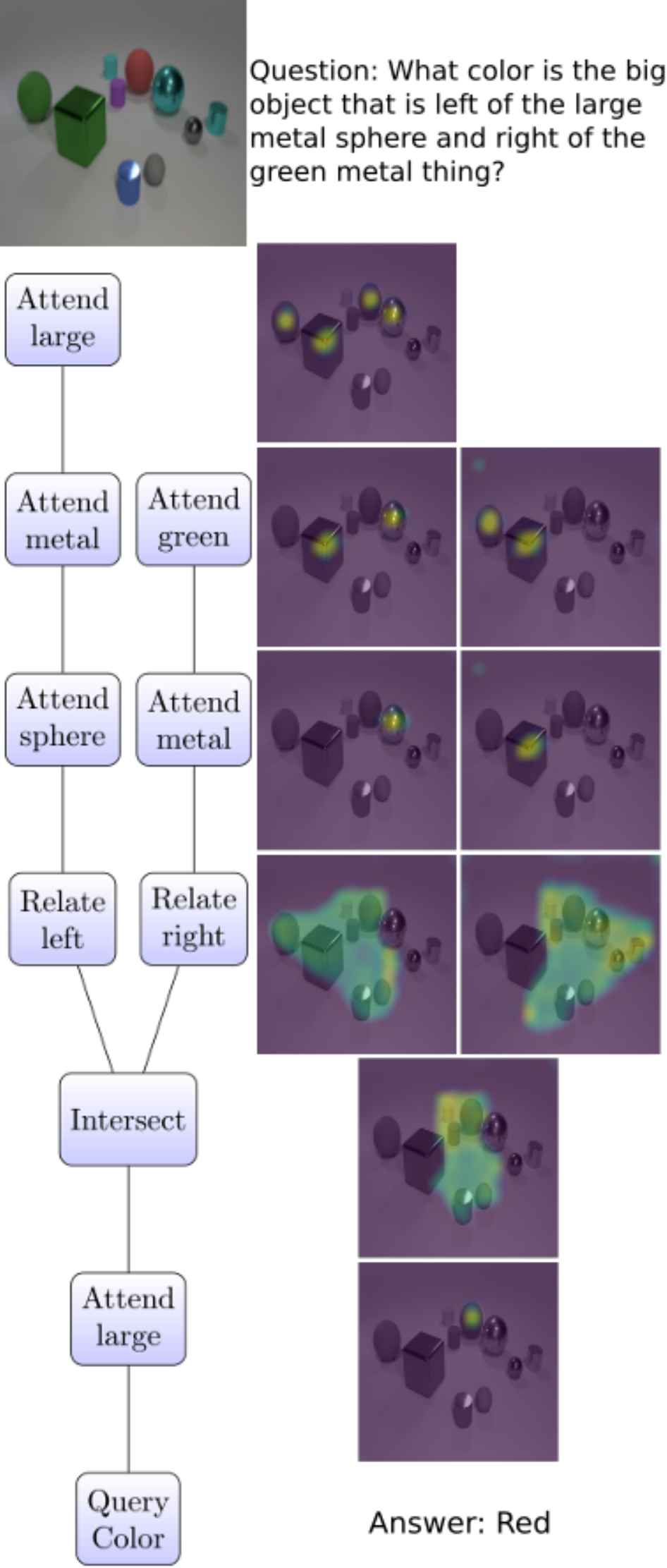}
  \caption{Read from top to bottom, a Transparency by Design network (TbD-net) composes visual
    attention masks to answer a question about objects in a scene. The tree diagram (left) indicates
    the modules used by the TbD-net, and their corresponding attention masks are depicted on the
    right.}
  \label{fig:attention-sensible}
\end{figure}

In the visual question answering task, most steps in the reasoning chain require localizing objects
that have some distinguishing visible property (\eg color, material, \etc). We ensure that each TbD
module performing this type of filtering outputs a one-dimensional attention mask, which explicitly
demarcates the relevant spatial regions. Thus, rather than refine high-dimensional feature maps
throughout the network, a TbD-net passes only attention masks between its modules. By intentionally
forcing this behavior, we produce a strikingly interpretable and intuitive model. This marks a step
away from complex neural networks as black boxes. Figure~\ref{fig:attention-sensible} shows an
example of how a TbD-net's attention shifts appropriately throughout its reasoning chain as it
solves a complex VQA problem, and that this process is easily interpretable via direct visualization
of the attention masks it produces. Note that our modules' use of attention is not a learnable
\emph{option}, as it is in the work of Hu \etal \cite{n2nmn}. Rather, our modules \emph{must}
utilize the attention that is passed into them, and thus must produce precise attention maps. All
the attention masks we display were generated using a perceptually-uniform color map
\cite{matplotlib}.

\subsection{Architecture Details}\label{sec:architecture}
We now describe the architecture of each of our modules. Table~\ref{tab:modules} provides an
overview of each module type. Several modules share input and output types (\eg \texttt{Attention}
and \texttt{Relate}) but differ in implementation, which is suited to their particular
task. Additional details on the implementation and operation of each module can be found in the
supplementary material.

\begin{table*}[!htb]
  \centering
  \caption[Neural Module Architectures]{A summary of the modules used in our Transparency by Design
    network. `Attention' and `Encoding' refer to single- and high-dimensional outputs, respectively,
    from a preceding module. `Stem' refers to image features produced by a trained neural
    network. The variables $x$ and $y$ refer to distinct objects in the scene, while [property]
    refers to one of color, shape, size, or material.}
  \label{tab:modules}
  \vspace{4pt}
  \centering
  \begin{tabular}{l l l}
    \toprule
    Module Type & Operation & Language Analogue\\\midrule
    \texttt{Attention} & Attention $\times$ Stem $\rightarrow$ Attention & Which things are [property]?\\
    \texttt{Query} & Attention $\times$ Stem $\rightarrow$ Encoding & What [property] is $x$?\\
    \texttt{Relate} & Attention $\times$ Stem $\rightarrow$ Attention & Left of, right of, in front, behind\\
    \texttt{Same} & Attention $\times$ Stem $\rightarrow$ Attention & Which things are the same [property] as $x$?\\
    \texttt{Comparison} & Encoding $\times$ Encoding $\rightarrow$ Encoding & Are $x$ and $y$ the same [property]?\\
    \texttt{And} & Attention $\times$ Attention $\rightarrow$ Attention & Left of $x$ \emph{and} right of $y$\\
    \texttt{Or} & Attention $\times$ Attention $\rightarrow$ Attention & Left of $x$ \emph{or} right of $y$\\
    \bottomrule
  \end{tabular}
\end{table*}

We use image features extracted from ResNet-101 \cite{resnet} and feed these through a simple
convolutional block called the `stem,' following the work of Johnson \etal \cite{clevr-iep}.
Similar to Hu \etal \cite{n2nmn}, and a point of departure from Johnson \etal \cite{clevr-iep} and
Andreas \etal \cite{nmn}, we provide stem features to most of our modules. This ensures image
features are readily accessible to each module and no information is lost in long compositions. The
stem translates high-dimensional feature input from ResNet into lower-dimensional features suitable
for our task.

\textbf{\texttt{Attention}} modules attend to the regions of the image that contain an object with a
specified property. For example, this type of module would be used to locate the red objects in a
scene. The \texttt{Attention} module takes as input image features from the stem and a previous
attention to refine (or an all-one tensor if it is the first \texttt{Attention} in the network) and
outputs a heatmap of dimension $1 \times H \times W$ corresponding to the objects of interest, which
we refer to as an \emph{attention mask}. This is done by multiplying the input image features by the
input attention mask elementwise, then processing those attended features with a series of
convolutions.

Logical \textbf{\texttt{And}} and \textbf{\texttt{Or}} modules combine two attention masks in a set
intersection and union, respectively. These operations need not be learned, since they are already
well-defined and can be implemented by hand. The \texttt{And} module takes the elementwise minimum
of two attention masks, spatially, while the \texttt{Or} module takes the elementwise maximum.

A \textbf{\texttt{Relate}} module attends to a region that has some spatial relation to another
region. For example, in the question ``What color is the cube to the right of the small sphere?'',
the network should determine the position of the small sphere using a series of \texttt{Attention}
modules, then use a \texttt{Relate} module to attend to the region that is spatially to the
right. This module needs global context in its operation, so that regions to the far right can be
influenced by an object on the far left, for instance. Zhu \etal \cite{structured-attentions} aptly
note that common VQA architectures have receptive fields that are too small for this global
information propagation. They propose a structured solution using a conditional random field. Our
solution is to use a series of dilated convolutions \cite{yu-koltun} in order to expand the
receptive field to the entire image, providing the global context needed by \texttt{Relate}. These
modules take as input image features from the stem and the attention mask from the previous module
and output an attention mask.

A \textbf{\texttt{Same}} module attends to a region, extracts a relevant property from that region,
and attends to every other region in the image that shares that property. As an example, when
answering the question ``Is anything the same color as the small cube?'', the network should
localize the small cube via \texttt{Attention} modules, then use a \texttt{Same} module to determine
its color and output an attention mask localizing all other objects sharing that color. As with the
\texttt{Relate} module, a \texttt{Same} module must take into account context from distant spatial
regions. However, the \texttt{Same} operation differs in its execution, since it must perform a more
complex function. We perform a cross-correlation between the object of interest and every other
object in the scene to determine which objects share the same property as the object of interest,
then send this output through a convolutional layer to produce an attention mask. Hence, the
\texttt{Same} modules take as input stem features and an attention mask and produce an attention
mask.

\textbf{\texttt{Query}} modules extract a feature from an attended region of an image. For example,
these modules would determine the color of an object of interest. Each \texttt{Query} takes as input
stem features and an attention mask and produces a feature map encoding the relevant property. The
image features are multiplied by the input attention elementwise, then processed by a series of
convolutions. A \texttt{Query} module is also used for determining whether a given description
matches any object (existence questions) and for counting.

A \textbf{\texttt{Compare}} module compares the properties output by two \texttt{Query} modules and
produces a feature map which encodes whether the properties are the same. This module would be used
to answer the question ``Are the cube and the metal sphere the same size?'', for example. Two
feature maps from \texttt{Query} modules are provided as input, concatenated, then processed by a
series of convolutions. A feature map is output, encoding the relevant information.

The final piece of our module network is a classifier that takes as input the feature map from
either a \texttt{Query} or \texttt{Compare} module and produces a distribution over answers. We
again follow the work of Johnson \etal \cite{clevr-iep}, using a series of convolutions, followed by
max-pooling and fully-connected layers.

\section{Experiments}\label{sec:experiments}
We evaluate our model using the CLEVR dataset \cite{clevr} and CLEVR-CoGenT. CLEVR is a VQA dataset
consisting of a training set of 70k images and 700k questions, as well as test and validation sets
of 15k images and 150k questions about objects in a rendered three-dimensional scene designed to
test compositional reasoning. For more details about the task, we refer the reader to Johnson \etal
\cite{clevr}.

In our model, all convolutional filters are initialized as described by He \etal \cite{msra}. Note
that our architectural changes do not affect the natural language processing component from Johnson
\etal \cite{clevr-iep}, which determines the composition of a modular network. For simplicity, we
use ground truth programs to train our network, because we do not modify the program generator. We
find that training on ground truth programs does not affect the accuracy of the model from Johnson
\etal \cite{clevr-iep}, compared to training with generated programs. Our training procedure thus
takes triplets $(x, z, a)$ of image, program, and answer for training. We use the Adam optimization
method \cite{adam} with learning rate set to $10^{-4}$ and our module network is trained end-to-end
with early stopping. The training procedures for CLEVR and for CoGenT are the same. Ground truth
programs are not provided with the CLEVR and CoGenT test sets, so we use the program generator from
Johnson \etal \cite{clevr-iep} to compute programs from questions. The testing procedure thus takes
image and question pairs $(x, q)$, produces a program $\pi(q)$ to arrange modules, then produces an
answer $\hat{a} = M_{\pi(q)}(x)$ where $M_{\pi(q)}$ is the arrangement of modules produced by the
program generator.

\subsection{CLEVR}\label{sec:performance}
Our initial model achieves 98.7\% test accuracy on the CLEVR dataset, far outperforming other neural
module network-based approaches. As we will describe, we utilize the attention masks produced by our
model to refine this initial model, resulting in state-of-the-art performance of 99.1\%
accuracy. Given the large number of high-performing models on CLEVR \cite{clevr-iep, ddrprog, cans,
  no-strong-priors}, we train our model 5 times for a statistical measure of performance, achieving
mean validation accuracy of 99.1\% with standard deviation 0.07. Further, we note that none of these
other models are amenable to having their reasoning processes inspected in an intuitive way. As we
will show, our model provides straightforward, interpretable outputs at \emph{every stage} of the
visual reasoning process.

\subsubsection{Attention as a Diagnostic Tool}\label{sec:diagnostic}
\begin{figure}[!htb]
  \includegraphics[keepaspectratio=true, width=\columnwidth]{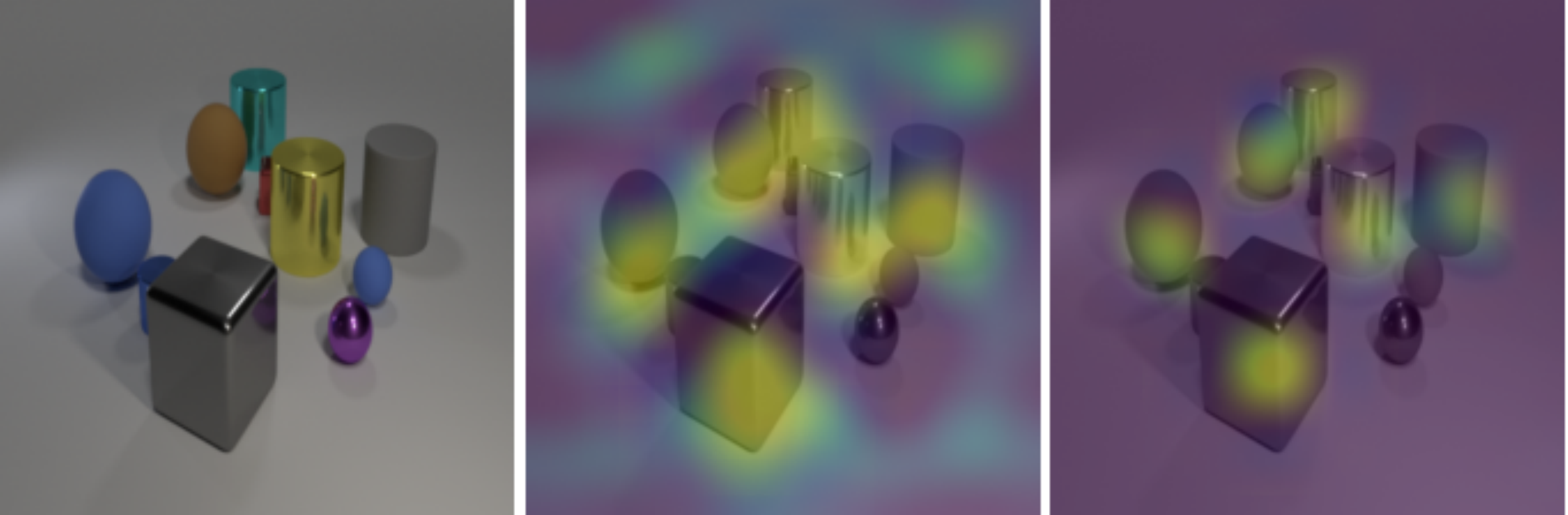}
  \caption{An input image (left) and the attention mask produced by the \texttt{Attention[large]}
    module overlaid atop the input image. Without penalizing attention mask outputs (middle) the
    attention mask is noisy and produces responses on background regions. Penalizing the attention
    outputs (right) provides a signal to reduce extraneous attention.}
  \label{fig:reg-compare}
\end{figure}

\textbf{Regularization}. Examining the attention masks produced by our initial model, we noticed
noise in the background. While not detrimental to our model's performance, these spurious regions of
attention may be confusing to users attempting to garner an understanding of the model outputs. Our
intuition behind this behavior is that the model is not penalized for producing small amounts of
attention on background regions because the later \texttt{Query} modules are able to effectively
ignore them, since they contain no objects. As a result, no error signal propagates back through the
model to push these to zero. To address this, we apply weighted $L_1$ regularization to the
intermediate attention mask outputs, providing an explicit signal to minimize unnecessary
activations. Experimentally, we find a factor of $2.5\times10^{-7}$ to be effective in reducing
spurious attention while maintaining strong attentions in true regions of interest. A comparison of
an \texttt{Attention} module output with and without this regularization can be seen in
Figure~\ref{fig:reg-compare}. Without regularization, the module produces a small amount of
attention on background regions, high attention on objects of interest, and zero attention on all
other objects. When we add this regularization term, the spurious background activations fade,
leaving a much more precise attention mask.

\begin{figure}[!b]
  \includegraphics[keepaspectratio=true, width=\columnwidth]{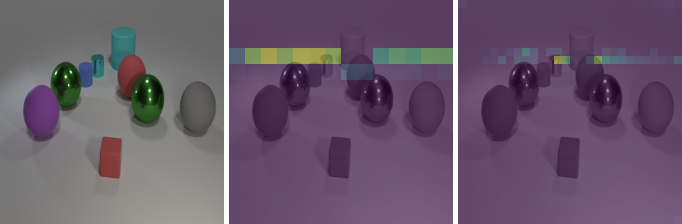}
  \caption{An input image (left) and the attention mask produced by the model when asked to attend
    to the region behind the blue rubber object and in front of the large cyan rubber cylinder with
    $14\times14$ (middle) and $28\times28$ (right) input features.}
  \label{fig:resolution-issue}
\end{figure}

\textbf{Spatial resolution}. Examining the attention masks from our initial model indicated the
resolution of the input feature maps is crucial for performance. Originally, our model took
$14\times14$ feature maps as input to each module. However, this was insufficient for resolving
closely-spaced objects. Increasing the resolution of the input feature maps to $28\times28$
alleviates these issues, which we do by extracting features from an earlier layer of
ResNet. Figure~\ref{fig:resolution-issue} shows our model operating over $14\times14$ and
$28\times28$ feature maps when asked to attend to an object in a narrow spatial region. Operating on
$14\times14$ feature maps, our model coarsely defines the region and incorrectly guesses that the
object is a sphere. Using $28\times28$ feature maps refines the region and allows the model to
correctly identify the cylinder.

\textbf{Performance improvements}. By adding regularization and increasing spatial resolution, two
strategies developed by examining our model attentions, we improve the accuracy of our model on
CLEVR from 98.7\% to 99.1\%, a new state of the art. A full table of results including comparisons
with several other approaches can be seen in Table~\ref{tab:clevr-acc}. While conceptually similar
to prior work in modular networks \cite{nmn, n2nmn, clevr-iep}, our model differs drastically in
module design. In addition to achieving superior performance, our paradigm allows for the
verification of module behavior and informs network design.

\begin{table*}[!htb]
  \centering
  \caption[CLEVR Dataset Accuracy]{Performance comparison of state-of-the-art models on the CLEVR
    dataset. Our model performs well while maintaining model transparency. We achieve state of the
    art performance on Query questions, while remaining competitive in all other categories. Our TbD
    model is trained without regularizing the output attention masks, while `+ reg' indicates the
    use of the regularization scheme described in the text. The `+ hres' indicator shows a model was
    trained using higher-resolution $28\times28$ feature maps rather than $14\times14$ feature
    maps.}
  \label{tab:clevr-acc}
  \vspace{4pt}
  \begin{tabular}{l c c c c c c}
    \toprule
    \multirow{2}{*}{Model} & \multirow{2}{*}{Overall} & \multirow{2}{*}{Count} & Compare & \multirow{2}{*}{Exist} & Query & Compare\\
    & & & Numbers & & Attribute & Attribute\\
    \midrule
    NMN \cite{nmn} & 72.1 & 52.5 & 72.7 & 79.3 & 79.0 & 78.0\\
    N2NMN \cite{n2nmn} & 88.8 & 68.5 & 84.9 & 85.7 & 90.0 & 88.8\\
    Human \cite{clevr} & 92.6 & 86.7 & 86.4 & 96.6 & 95.0 & 96.0\\
    CNN + LSTM + RN \cite{relational-reasoning} & 95.5 & 90.1 & 93.6 & 97.8 & 97.1 & 97.9\\
    PG + EE (700k) \cite{clevr-iep} & 96.9 & 92.7 & 98.7 & 97.1 & 98.1 & 98.9\\
    CNN + GRU + FiLM \cite{no-strong-priors} & 97.6 & 94.5 & 93.8 & 99.2 & 99.2 & 99.0\\
    DDRprog \cite{ddrprog} & 98.3 & 96.5 & 98.4 & 98.8 & 99.1 & 99.0\\
    MAC \cite{cans} & 98.9 & 97.2 & \textbf{99.4} & \textbf{99.5} & 99.3 & 99.5\\
    \midrule
    TbD-net (Ours) & 98.7 & 96.8 & 99.1 & 98.9 & 99.4 & 99.2\\
    TbD + reg (Ours) & 98.5 & 96.5 & 99.0 & 98.9 & 99.3 & 99.1\\
    TbD + reg + hres (Ours) & \textbf{99.1} & \textbf{97.6} & \textbf{99.4} & 99.2 & \textbf{99.5} & \textbf{99.6}\\
    \bottomrule
  \end{tabular}
\end{table*}

\subsubsection{Transparency}\label{sec:transparency}
We examine the attention masks produced by the intermediate modules of our TbD model. We show that
our model explicitly composes visual attention masks to arrive at an answer, leading to an
unprecedented level of transparency in the neural network. Figure~\ref{fig:attention-sensible} shows
the composition of visual attentions through an entire question. In this section, we provide a
quantitative analysis of transparency. We further examine the outputs of several modules, displayed
without any smoothing, showing that each step of any composition is straightforwardly interpretable.

\textbf{Quantitative Analysis of Attention}. Here we propose a quantitative analysis of the
interpretability of visual attention, which will allow for direct comparison with subsequent work in
this area. In this context, a module's attention is interpretable if it visually highlights the
correct objects in a scene, without ambiguity. Specifically, we measure how often the center-of-mass
of an attended region overlaps with the appropriate regions of the ground truth segmentation. We
perform this analysis for each of our \texttt{Attention} modules within a given chain of
reasoning. The CLEVR dataset generator \cite{clevr} was used to produce 1k images and 10k questions
for evaluation. This analysis produces precision and recall metrics to summarize the interpretable
quality of a model's attention.

Our original TbD-net model has a recall of 0.86 and a precision of 0.41. This low
precision is largely due to attention placed on the background (analyzing the performance on
foreground objects alone yields a precision of 0.95). The modifications to our model detailed in
Section~\ref{sec:diagnostic} dramatically improve the interpretability metrics as evaluated on the
full image (foreground and background). Adding regularization improves the recall and precision
values to 0.92 and 0.90, respectively, and increasing the spatial resolution further improves the
values to 0.99 and 0.98, respectively.

\begin{figure}[!t]
  \centering \includegraphics[keepaspectratio=true,
    width=\columnwidth]{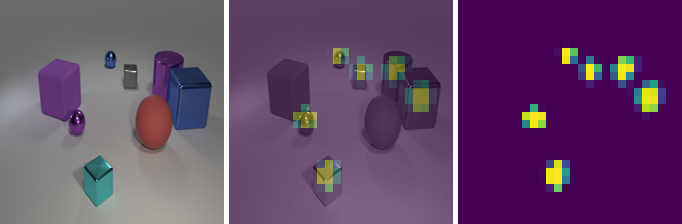}
  \caption{An input image (left) and the attention mask produced by the \texttt{Attention[metal]}
    module (right). When the attention mask is overlaid atop the input image (middle), it is
    apparent that the attention is appropriately focused on the metal objects.}
  \label{fig:im000021-metal}
\end{figure}

\textbf{Qualitative Analysis of Attention}. Figure~\ref{fig:im000021-metal} shows the output of an
\texttt{Attention} module that focuses on metal objects. The output is sensible and matches our
intuition of where attention should be placed in the image. Maximal attention is given to the metal
objects and minimal attention to the rubber objects and to the background region.

\begin{figure}[!b]
  \centering \includegraphics[keepaspectratio=true,
    width=\columnwidth]{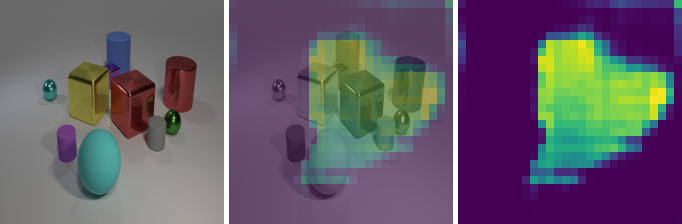}
  \caption{An input image (left) and the attention mask produced by the \texttt{Relate[right]}
    module (right) when it receives an attention on the purple cylinder. When the attention mask is
    overlaid atop the input image (middle), it is apparent that the attention is focused on the
    region to the right of the purple cylinder.}
  \label{fig:relate}
\end{figure}

\begin{figure}[!h]
  \centering
  \includegraphics[keepaspectratio=true, width=\columnwidth]{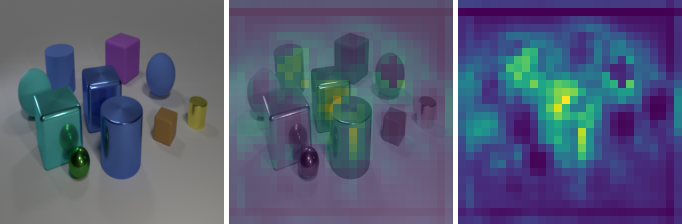}
  \caption{An input image (left) and the attention mask produced by the \texttt{Same[color]} module
    (right) when it receives an attention on the blue sphere. The attention mask overlaid atop the
    input image (middle) demonstrates that the module successfully performs the complex operation
    of (1) determining the color of the sphere (2) determining the color of all the other objects in
    the scene and (3) attending to the objects with the same color.}
  \label{fig:same}
\end{figure}

The \texttt{Attention} modules are the simplest of the visual primitives. However, more complex
operations, such as \texttt{Same} and \texttt{Relate}, still produce intuitive attention masks. We
show in Figures~\ref{fig:relate} and \ref{fig:same} that these modules are just as transparent and
easy to understand. The \texttt{Relate} module is given an attention mask highlighting the purple
cylinder and shifts attention to the right. We see it highlights the entire region of the image,
which it is able to do due to its expanded receptive field. The \texttt{Same} module is given an
attention mask focusing on the blue sphere, then is asked to shift its focus to the objects of the
same color. It gives maximal attention to the other three blue objects, minimal attention to all
other objects, and a small amount of attention to the background.

Our logical \texttt{And} and \texttt{Or} modules perform set intersection and union operations,
respectively. Their behaviors are as intuitive as those set operations. Since \texttt{Query} and
\texttt{Compare} modules output feature maps rather than attention masks, their outputs cannot be
easily visualized. This is not problematic, as their corresponding operations do not have clear
visual anchors. That is, it is not sensible to highlight image regions based on concepts such as
`what shape' or `are the same number,' which have no referent without additional textual context.

\subsection{CLEVR-CoGenT}
The CLEVR-CoGenT dataset provides an excellent test for generalization. It is identical in form to
the CLEVR dataset with the exception that it has two different conditions. In Condition A all cubes
are colored one of gray, blue, brown, or yellow, and all cylinders are one of red, green, purple, or
cyan; in Condition B the color palettes are swapped. This provides a check that the model does not
tie together the notions of shape and color. In this section, we report only our best-performing
model. Like previous work \cite{clevr-iep}, our performance is worse on Condition B than Condition A
after training only using Condition A data. As Table~\ref{tab:cogent-acc} shows, our model achieves
98.8\% accuracy on Condition A, but only 75.4\% on Condition B. Following Johnson \etal
\cite{clevr-iep}, we then fine-tune our model using 3k images and 30k questions from the Condition B
data. Whereas other models see a significant drop in performance on the Condition A data after
fine-tuning, our model maintains high performance. As Table~\ref{tab:cogent-acc} shows, our model
can effectively learn from a small amount of Condition B data. We achieve 96.9\% accuracy on
Condition A and 96.3\% accuracy on Condition B after fine-tuning, far surpassing the
highest-reported 76.1\% Condition A and 92.7\% Condition B accuracy.

\begin{table}[!tb]
  \centering
  \caption{Performance comparison against the current state-of-the-art model on the CoGenT dataset
    having trained only on Condition A data (middle column) and after fine-tuning on a small amount
    of data with novel attributes (right column).}
  \label{tab:cogent-acc}
  \vspace{4pt}
  \begin{tabular}{l c c c c}
    \toprule
    & \multicolumn{2}{c}{Train A} & \multicolumn{2}{c}{Fine-tune B}\\
    \addlinespace[-2pt]\cmidrule(lr){2-3}\cmidrule(lr){4-5}\addlinespace[-2pt]
     & A & B & A & B\\
    \addlinespace[-2pt]\midrule
    PG + EE \cite{clevr-iep} & 96.6 & 73.7 & 76.1 & 92.7\\
    TbD + reg (Ours) & \textbf{98.8} & \textbf{75.4} & \textbf{96.9} & \textbf{96.3}\\
    \bottomrule
  \end{tabular}
\end{table}

We perform an analysis of conditional probabilities based on shape/color dependencies to determine
the cause of our model's poor performance on Condition B before fine-tuning. Using the dataset from
Section~\ref{sec:transparency}, we demonstrate that the model's ability to identify shape, in
particular, depends heavily on the co-occurrence of shape and color. Table~\ref{tab:entangled} shows
that the model is only able to effectively identify shapes in colors it has seen before fine-tuning,
while it can identify color regardless of shape. Fine-tuning on Condition B rectifies the
entanglement.

\begin{table}[!tb]
  \setlength{\tabcolsep}{4pt}
  \centering
  \caption{Our model's ability to determine the shape of an object depends heavily on the
    co-occurrence of shape and color before fine-tuning, while its ability to determine the color of
    an object does not. $P(\checkmark)$ indicates the probability of correctly identifying an
    object. $A$ and $B$ correspond to the CoGenT color/shape splits.}
  \label{tab:entangled}
  \vspace{4pt}
  \begin{tabular}{l c c c c}
    \toprule
     & \multicolumn{2}{c}{Predict Shape} & \multicolumn{2}{c}{Predict Color}\\
    \addlinespace[-1pt]\cmidrule(lr){2-3}\cmidrule(lr){4-5}\addlinespace[-2pt]
    & $P(\checkmark|A)$ & $P(\checkmark|B)$ & $P(\checkmark|A)$ & $P(\checkmark|B)$\\
    \midrule
    Train A & 0.90 & 0.22 & 0.91 & 0.84\\
    Fine-tune B & 0.77 & 0.81 & 0.90 & 0.86\\
    \bottomrule
  \end{tabular}
\end{table}

\begin{figure}[!tb]
  \centering
  \includegraphics[keepaspectratio=true, width=\columnwidth]{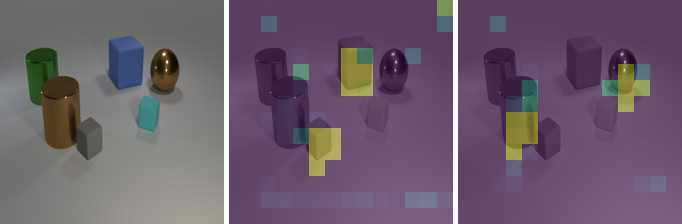}
  \caption{An input image (left) and the overlaid attention masks produced by
    \texttt{Attention[cube]} (middle) and \texttt{Attention[brown]} (right). Having never seen a
    cyan cube, \texttt{Attention[cube]} misses that object. Having never seen a brown cylinder, the
    model mistakenly places some attention on the brown cylinder in the scene. Conversely,
    \texttt{Attention[brown]} generalizes correctly, regardless of shape.}
  \label{fig:cogent}
\end{figure}

Figure~\ref{fig:cogent} shows this entanglement visually. When asked to attend to cubes before
fine-tuning, our model correctly focuses on the gray and blue cubes in that image, but ignores the
cyan cube and incorrectly places some attention on the brown cylinder. What is particularly
noteworthy is the fact that our model's representations of color are complete (with respect to
CLEVR). When the network is asked to attend to the brown objects, it correctly identifies both the
sphere and cylinder, even though it has only seen cubes and spheres in brown. Thus, our model has
entangled its representation of shape with color, but has not entangled its representation of color
with shape.

\section{Discussion}\label{sec:discussion}
We have presented Transparency by Design networks, which compose visual primitives that leverage an
explicit attention mechanism to perform reasoning operations. Unlike their predecessors, the
resulting neural module networks are both highly performant and readily interpretable. This is a key
advantage to utilizing TbD models\textemdash the ability to directly evaluate the model's learning
process via the produced attention masks is a powerful diagnostic tool. One can leverage this
capability to inspect the semantics of a visual operation, such as `same color,' and redesign
modules to address apparent aberrations in reasoning. Using these attentions as a means to improve
performance, we achieve state-of-the-art accuracy on the challenging CLEVR dataset and the CoGenT
generalization task. Such insight into a neural network's operation may also help build user trust
in visual reasoning systems.

{ \small \bibliographystyle{ieee} \bibliography{tbd-nets} }

\begin{thebibliography}{10}\itemsep=-1pt

\bibitem{bottom-up-top-down}
P.~{Anderson}, X.~{He}, C.~{Buehler}, D.~{Teney}, M.~{Johnson}, S.~{Gould}, and
  L.~{Zhang}.
\newblock {Bottom-Up and Top-Down Attention for Image Captioning and VQA}.
\newblock {\em ArXiv e-prints}, July 2017.

\bibitem{nmn}
J.~Andreas, M.~Rohrbach, T.~Darrell, and D.~Klein.
\newblock Deep compositional question answering with neural module networks.
\newblock {\em CoRR}, abs/1511.02799, 2015.

\bibitem{learning-to-compose}
J.~Andreas, M.~Rohrbach, T.~Darrell, and D.~Klein.
\newblock Learning to compose neural networks for question answering.
\newblock {\em CoRR}, abs/1601.01705, 2016.

\bibitem{abc-cnn}
K.~Chen, J.~Wang, L.~Chen, H.~Gao, W.~Xu, and R.~Nevatia.
\newblock {ABC-CNN:} an attention based convolutional neural network for visual
  question answering.
\newblock {\em CoRR}, abs/1511.05960, 2015.

\bibitem{cans}
C.~D.~M. Drew Arad~Hudson.
\newblock Compositional attention networks for machine reasoning.
\newblock {\em International Conference on Learning Representations}, 2018.
\newblock accepted as poster.

\bibitem{mcb}
A.~{Fukui}, D.~{Huk Park}, D.~{Yang}, A.~{Rohrbach}, T.~{Darrell}, and
  M.~{Rohrbach}.
\newblock {Multimodal Compact Bilinear Pooling for Visual Question Answering
  and Visual Grounding}.
\newblock {\em ArXiv e-prints}, June 2016.

\bibitem{segmentation-aware}
A.~W. Harley, K.~G. Derpanis, and I.~Kokkinos.
\newblock Segmentation-aware convolutional networks using local attention
  masks.
\newblock {\em CoRR}, abs/1708.04607, 2017.

\bibitem{resnet}
K.~He, X.~Zhang, S.~Ren, and J.~Sun.
\newblock Deep residual learning for image recognition.
\newblock {\em CoRR}, abs/1512.03385, 2015.

\bibitem{msra}
K.~He, X.~Zhang, S.~Ren, and J.~Sun.
\newblock Delving deep into rectifiers: Surpassing human-level performance on
  imagenet classification.
\newblock {\em CoRR}, abs/1502.01852, 2015.

\bibitem{n2nmn}
R.~Hu, J.~Andreas, M.~Rohrbach, T.~Darrell, and K.~Saenko.
\newblock Learning to reason: End-to-end module networks for visual question
  answering.
\newblock {\em CoRR}, abs/1704.05526, 2017.

\bibitem{matplotlib}
J.~D. Hunter.
\newblock Matplotlib: A 2d graphics environment.
\newblock {\em Computing In Science \& Engineering}, 9(3):90--95, 2007.

\bibitem{video-summarization}
Z.~Ji, K.~Xiong, Y.~Pang, and X.~Li.
\newblock Video summarization with attention-based encoder-decoder networks.
\newblock {\em CoRR}, abs/1708.09545, 2017.

\bibitem{clevr}
J.~Johnson, B.~Hariharan, L.~van~der Maaten, L.~Fei{-}Fei, C.~L. Zitnick, and
  R.~B. Girshick.
\newblock {CLEVR:} {A} diagnostic dataset for compositional language and
  elementary visual reasoning.
\newblock {\em CoRR}, abs/1612.06890, 2016.

\bibitem{clevr-iep}
J.~Johnson, B.~Hariharan, L.~van~der Maaten, J.~Hoffman, L.~Fei-Fei, C.~L.
  Zitnick, and R.~Girshick.
\newblock Inferring and executing programs for visual reasoning.
\newblock In {\em ICCV}, 2017.

\bibitem{adam}
D.~P. Kingma and J.~Ba.
\newblock Adam: {A} method for stochastic optimization.
\newblock {\em CoRR}, abs/1412.6980, 2014.

\bibitem{mythos-interpretability}
Z.~C. {Lipton}.
\newblock {The Mythos of Model Interpretability}.
\newblock {\em ArXiv e-prints}, June 2016.

\bibitem{picanet}
N.~Liu and J.~Han.
\newblock Picanet: Learning pixel-wise contextual attention in convnets and its
  application in saliency detection.
\newblock {\em CoRR}, abs/1708.06433, 2017.

\bibitem{hierarchical-coattention}
J.~{Lu}, J.~{Yang}, D.~{Batra}, and D.~{Parikh}.
\newblock {Hierarchical Question-Image Co-Attention for Visual Question
  Answering}.
\newblock {\em ArXiv e-prints}, May 2016.

\bibitem{no-strong-priors}
E.~{Perez}, H.~{de Vries}, F.~{Strub}, V.~{Dumoulin}, and A.~{Courville}.
\newblock {Learning Visual Reasoning Without Strong Priors}.
\newblock {\em ArXiv e-prints}, July 2017.

\bibitem{person-reid}
A.~{Rahimpour}, L.~{Liu}, A.~{Taalimi}, Y.~{Song}, and H.~{Qi}.
\newblock {Person Re-identification Using Visual Attention}.
\newblock {\em ArXiv e-prints}, July 2017.

\bibitem{relational-reasoning}
A.~{Santoro}, D.~{Raposo}, D.~G.~T. {Barrett}, M.~{Malinowski}, R.~{Pascanu},
  P.~{Battaglia}, and T.~{Lillicrap}.
\newblock {A simple neural network module for relational reasoning}.
\newblock {\em ArXiv e-prints}, June 2017.

\bibitem{gradcam}
R.~R. {Selvaraju}, M.~{Cogswell}, A.~{Das}, R.~{Vedantam}, D.~{Parikh}, and
  D.~{Batra}.
\newblock {Grad-CAM: Visual Explanations from Deep Networks via Gradient-based
  Localization}.
\newblock {\em ArXiv e-prints}, Oct. 2016.

\bibitem{where-to-look}
K.~J. {Shih}, S.~{Singh}, and D.~{Hoiem}.
\newblock {Where To Look: Focus Regions for Visual Question Answering}.
\newblock {\em ArXiv e-prints}, Nov. 2015.

\bibitem{ddrprog}
J.~Suarez, J.~Johnson, and F.-F. Li.
\newblock {DDR}prog: A {CLEVR} differentiable dynamic reasoning programmer,
  2018.

\bibitem{residual-attention}
F.~{Wang}, M.~{Jiang}, C.~{Qian}, S.~{Yang}, C.~{Li}, H.~{Zhang}, X.~{Wang},
  and X.~{Tang}.
\newblock {Residual Attention Network for Image Classification}.
\newblock {\em ArXiv e-prints}, Apr. 2017.

\bibitem{ask-attend-answer}
H.~{Xu} and K.~{Saenko}.
\newblock {Ask, Attend and Answer: Exploring Question-Guided Spatial Attention
  for Visual Question Answering}.
\newblock {\em ArXiv e-prints}, Nov. 2015.

\bibitem{show-attend-tell}
K.~Xu, J.~Ba, R.~Kiros, K.~Cho, A.~C. Courville, R.~Salakhutdinov, R.~S. Zemel,
  and Y.~Bengio.
\newblock Show, attend and tell: Neural image caption generation with visual
  attention.
\newblock {\em CoRR}, abs/1502.03044, 2015.

\bibitem{sans}
Z.~{Yang}, X.~{He}, J.~{Gao}, L.~{Deng}, and A.~{Smola}.
\newblock {Stacked Attention Networks for Image Question Answering}.
\newblock {\em ArXiv e-prints}, Nov. 2015.

\bibitem{yu-koltun}
F.~Yu and V.~Koltun.
\newblock Multi-scale context aggregation by dilated convolutions.
\newblock In {\em ICLR}, 2016.

\bibitem{mcb-co}
Z.~{Yu}, J.~{Yu}, J.~{Fan}, and D.~{Tao}.
\newblock {Multi-modal Factorized Bilinear Pooling with Co-Attention Learning
  for Visual Question Answering}.
\newblock {\em ArXiv e-prints}, Aug. 2017.

\bibitem{zhang2018learning}
Y.~Zhang, J.~Hare, and A.~Prügel-Bennett.
\newblock Learning to count objects in natural images for visual question
  answering.
\newblock In {\em International Conference on Learning Representations}, 2018.

\bibitem{structured-attentions}
C.~{Zhu}, Y.~{Zhao}, S.~{Huang}, K.~{Tu}, and Y.~{Ma}.
\newblock {Structured Attentions for Visual Question Answering}.
\newblock {\em ArXiv e-prints}, Aug. 2017.

\bibitem{visual7w}
Y.~{Zhu}, O.~{Groth}, M.~{Bernstein}, and L.~{Fei-Fei}.
\newblock {Visual7W: Grounded Question Answering in Images}.
\newblock {\em ArXiv e-prints}, Nov. 2015.

\end{thebibliography}

\section*{Supplementary Material}

\setlength\tabcolsep{2pt}

\section{Module Details}
Here we describe each module in detail and provide motivation for specific architectural choices. In
all descriptions, `image features' refer to features that have been extracted using a pretrained
model \cite{resnet} and passed through our stem network, which is shared across modules. In all the
following tables, $\delta(\cdot)$ will indicate a rectified linear function and $\sigma(\cdot)$ will
indicate a sigmoid activation. The input size $R \times C$ indicates $R$ rows and $C$ columns in the
input. In our original model, $R = C = 14$, while our high-resolution model uses $R = C = 28$.

The architecture of the \textbf{\texttt{Attention}} modules can be seen in
Table~\ref{tab:attention-architecture}. These modules take stem features and an attention mask as
input and produce an attention mask as output. We first perform an elementwise multiplication of the
input features and attention mask, broadcasting the attention mask along the channel dimension of
the input features. We refer to this process as `attending to' the features. We then process the
attended features with two 3x3 convolutions, each with 128 filters and a ReLU, then use a single 1x1
convolution followed by a sigmoid to project down to an attention mask. This architecture is
motivated by the design of the unary module from Johnson \etal \cite{clevr-iep}.

\begin{table}[!b]
    \caption[\texttt{Attention} Module Architecture]{Architecture of an \texttt{Attention} module,
      which takes as input Features and an Attention and produces an Attention.}
    \label{tab:attention-architecture}
    \vspace{4pt}
    \centering
    \begin{tabular}{l l l}
      \toprule
        Index & Layer & Output Size\\\midrule
        (1) & Features & $128 \times R \times C$\\
        (2) & Previous module output & $1 \times R \times C$\\
        (3) & Elementwise multiply (1) and (2) & $128 \times R \times C$\\
        (4) & $\delta$(Conv($3 \times 3, 128 \rightarrow 128$)) & $128 \times R \times C$\\
        (5) & $\delta$(Conv($3 \times 3, 128 \rightarrow 128$)) & $128 \times R \times C$\\
        (6) & $\sigma$(Conv($1 \times 1, 128 \rightarrow 1$)) & $1 \times R \times C$\\
        \bottomrule
    \end{tabular}
\end{table}

The \textbf{\texttt{And}} and \textbf{\texttt{Or}} modules, seen in Table~\ref{tab:and-architecture}
and Table~\ref{tab:or-architecture}, respectively, perform set intersection and union
operations. These modules return the elementwise minimum and maximum, respectively, of two input
attention masks. This is motivated by the logical operations that Hu \etal \cite{n2nmn} implement,
which seems a natural expression of these operations. Such simple and straightforward operations
need not be learned, since they can be effectively and efficiently implemented by hand.

\begin{table}[!tb]
  \caption[\texttt{And} Module Architecture]{Architecture of an \texttt{And} module. This module
    receives as input two Attentions and produces an Attention.}
  \label{tab:and-architecture}
  \vspace{4pt}
  \centering
  \begin{tabular}{l l l}
    \toprule
    Index & Layer & Output Size\\\midrule
    (1) & Previous module output & $1 \times R \times C$\\
    (2) & Previous module output & $1 \times R \times C$\\
    (3) & Elementwise minimum (1) and (2) & $1 \times R \times C$\\
    \bottomrule
  \end{tabular}
\end{table}

\begin{table}[!tb]
  \caption[\texttt{Or} Module Architecture]{Architecture of an \texttt{Or} module. This module
    receives as input two Attentions and produces an Attention.}
  \label{tab:or-architecture}
  \vspace{4pt}
  \centering
  \begin{tabular}{l l l}
    \toprule
    Index & Layer & Output Size\\\midrule
    (1) & Previous module output & $1 \times R \times C$\\
    (2) & Previous module output & $1 \times R \times C$\\
    (3) & Elementwise maximum (1) and (2) & $1 \times R \times C$\\
    \bottomrule
  \end{tabular}
\end{table}

The \texttt{\textbf{Relate}} module, shown in Table~\ref{tab:relate-architecture}, needs global
context to shift attention across an entire image. Motivated by this, we use a series of dilated 3x3
convolutions, with dilation factors 1, 2, 4, 8, and 1, to expand the receptive field to the entire
image. The choice of dilation factors is informed by the work of Yu and Koltun
\cite{yu-koltun}. These modules receive stem features and an attention mask and produce an attention
mask. Each convolution in the series has 128 filters and is followed by a ReLU. A final convolution
then reduces the feature map to a single-channel attention mask, and a sigmoid nonlinearity is
applied.

\begin{table}[!tb]
  \caption[\texttt{Relate} Module Architecture]{Architecture of a \texttt{Relate} module. These
    modules receive as input Features and an Attention and produce an Attention.}
  \label{tab:relate-architecture}
  \vspace{4pt}
  \centering
  \begin{tabular}{l l l}
    \toprule
    Index & Layer & Output Size\\\midrule
    (1) & Features & $128 \times R \times C$\\
    (2) & Previous module output & $1 \times R \times C$\\
    (3) & Elementwise multiply (1) and (2) & $128 \times R \times C$\\
    (4) & $\delta$(Conv($3 \times 3, 128 \rightarrow 128$, dilate 1)) & $128 \times R \times C$\\
    (5) & $\delta$(Conv($3 \times 3, 128 \rightarrow 128$, dilate 2)) & $128 \times R \times C$\\
    (6) & $\delta$(Conv($3 \times 3, 128 \rightarrow 128$, dilate 4)) & $128 \times R \times C$\\
    (7) & $\delta$(Conv($3 \times 3, 128 \rightarrow 128$, dilate 8)) & $128 \times R \times C$\\
    (8) & $\delta$(Conv($3 \times 3, 128 \rightarrow 128$, dilate 1)) & $128 \times R \times C$\\
    (9) & $\sigma$(Conv($1 \times 1, 128 \rightarrow 1$)) & $1 \times R \times C$\\
    \bottomrule
  \end{tabular}
\end{table}

The \texttt{\textbf{Same}} module is the most complex of our modules, and the most complex operation
we perform. To illustrate this, consider the \texttt{Same[shape]} module. It must determine the
shape of the attended object, compare that shape with the shape of every other object in the scene
(which requires global information propagation), and attend to all the objects that share that
shape.  Initially, we used a design similar to the \texttt{Relate} module to perform this operation,
but found it did not perform well. After further reflection, we posited this was because the
\texttt{Relate} module does not have a mechanism for remembering which object we are interested in
performing the \texttt{Same} with respect to. Table~\ref{tab:same-architecture} provides an overview
of the \texttt{Same} module. Here we explicate the notation. Provided with stem features and an
attention mask as input, we take the $\arg\max$ of the feature map, spatially. This gives us the
$(x,y)$ position of the object of interest (\ie the object to perform the \texttt{Same} with respect
to). We then extract the feature vector at this spatial location in the input feature map, which
gives us the vector encoding the property of interest (among other properties). Next, we perform an
elementwise multiplication with the feature vector at every spatial dimension. This essentially
performs a cross-correlation of the feature vector of interest with the feature vector of every
other position in the image. Our intuition is that the vector dimensions that encode the property of
interest will be `hot' at every point sharing that property with the object of interest.  At this
point, a convolution could be learned that attends to the relevant regions. However, the
\texttt{Same} operation, by its definition in CLEVR, must \emph{not} attend to the original
object. That is, an object is by definition not the same property as itself. Therefore, the
\texttt{Same} module must learn not to attend to the original object. We thus concatenate the
original input attention mask with the cross-correlated feature map, allowing the convolutional
filter to know which object was the original, and thus ignore it.

\begin{table}[h]
  \caption[\texttt{Same} Module Architecture]{Architecture of a \texttt{Same} module. These
    modules receive as input Features and an Attention and produce an Attention.}
  \label{tab:same-architecture}
  \vspace{4pt}
  \centering
  \begin{tabular}{l l l}
    \toprule
    Index & Layer & Output Size\\\midrule
    (1) & Features & $128 \times R \times C$\\
    (2) & Previous module output & $1 \times R \times C$\\
    (3) & $\arg\max_{x,y}(2)$ & $1 \times 1 \times 1$\\
    (4) & $(1)_{(3)}$ & $128 \times 1 \times 1$\\
    (5) & Elementwise multiply (1) and (4) & $128 \times R \times C$\\
    (6) & Concatenate (5) and (2) & $129 \times R \times C$\\
    (7) & $\sigma$(Conv($1 \times 1, 129 \rightarrow 1$)) & $1 \times R \times C$\\
    \bottomrule
  \end{tabular}
\end{table}

The \texttt{\textbf{Query}} module architecture can be seen in Table~\ref{tab:query}. Its design is
similar to that of the \texttt{Attention} modules and is likewise inspired by the unary module
design of Johnson \etal \cite{clevr-iep} and adapted for receiving an attention mask and stem
features as input. These modules produce a feature map as output, and thus do not have a
convolutional filter that performs a down-projection.

\begin{table}[!tb]
  \caption[\texttt{Query} Module Architecture]{Architecture of a \texttt{Query} module. These
    modules receive as input Features and an Attention and produce an Encoding.}
  \label{tab:query}
  \vspace{4pt}
  \centering
  \begin{tabular}{l l l}
    \toprule
    Index & Layer & Output Size\\\midrule
    (1) & Features & $128 \times R \times C$\\
    (2) & Previous module output & $1 \times R \times C$\\
    (3) & Elementwise multiply (1) and (2) & $128 \times R \times C$\\
    (4) & $\delta$(Conv($3 \times 3, 128 \rightarrow 128$)) & $128 \times R \times C$\\
    (5) & $\delta$(Conv($3 \times 3, 128 \rightarrow 128$)) & $128 \times R \times C$\\
    \bottomrule
  \end{tabular}
\end{table}

The \texttt{\textbf{Compare}} module, shown in Table~\ref{tab:compare-architecture}, is inspired by
the binary module of Johnson \etal \cite{clevr-iep}. These modules take two feature maps as input
and produce a feature map as output. Their purpose is to determine whether the two input feature
maps encode the same property.

\begin{table}[!tb]
  \caption[\texttt{Compare} Module Architecture]{Architecture of a \texttt{Compare} module. These
    modules receive as input two Features and produce an Encoding.}
  \label{tab:compare-architecture}
  \vspace{4pt}
  \centering
  \begin{tabular}{l l l}
    \toprule
    Index & Layer & Output Size\\\midrule
    (1) & Previous module output & $128 \times R \times C$\\
    (2) & Previous module output & $128 \times R \times C$\\
    (3) & Concatenate (1) and (2) & $128 \times R \times C$\\
    (4) & $\delta$(Conv($1 \times 1, 128 \rightarrow 128$)) & $128 \times R \times C$\\
    (5) & $\delta$(Conv($3 \times 3, 128 \rightarrow 128$)) & $128 \times R \times C$\\
    (6) & $\delta$(Conv($3 \times 3, 128 \rightarrow 128$)) & $128 \times R \times C$\\
    \bottomrule
  \end{tabular}
\end{table}


\end{document}